\documentclass[conference]{IEEEtran}
\IEEEoverridecommandlockouts
\usepackage{cite}
\usepackage{amsmath,amssymb,amsfonts}
\usepackage{algorithmic}
\usepackage{graphicx}
\usepackage{textcomp}
\usepackage[dvipsnames]{xcolor} % dvipsnames for more color names.
\usepackage[colorlinks,bookmarks=false]{hyperref} % colorlinks, hidelinks
\usepackage{cleveref}
\usepackage{booktabs}
\usepackage{csquotes}
\MakeOuterQuote{"}

\ifCLASSOPTIONcompsoc
  \usepackage[caption=false,font=normalsize,labelfont=sf,textfont=sf]{subfig}
\else
  \usepackage[caption=false,font=footnotesize]{subfig}
\fi

\usepackage{makecell}
\usepackage{multirow}

\usepackage[utf8]{inputenc} % Usa UTF-8
\usepackage[T1]{fontenc} % https://tex.stackexchange.com/q/664

\def\BibTeX{{\rm B\kern-.05em{\sc i\kern-.025em b}\kern-.08em
    T\kern-.1667em\lower.7ex\hbox{E}\kern-.125emX}}

\usepackage{soul}
\setstcolor{red}
\setul{0.1pt}{1pt}

\newcommand{\mergeh}[3]{\begin{tabular}[#1]{@{}#2@{}}#3\end{tabular}}

\begin{document}

\title{Traffic Light Recognition Using Deep Learning and Prior Maps for Autonomous Cars
	% \thanks{This study was financed in part by the Coordenação de Aperfeiçoamento de Pessoal de Nível Superior -- Brasil (CAPES) -- Finance Code 001; by Conselho Nacional de Desenvolvimento Científico e Tecnológico (CNPq, Brazil) with scholarships of Productivity on Research (grants 311120/2016-4 and 311504/2017-5); and Fundação de Amparo à Pesquisa do Espírito Santo -- Brazil (FAPES) -- grant 75537958/16.}
  \thanks{This study was financed in part by Coordenação de Aperfeiçoamento de Pessoal de Nível Superior -- Brasil (CAPES) -- Finance Code 001; Conselho Nacional de Desenvolvimento Científico e Tecnológico -- Brasil (CNPq) -- grants 311120/2016-4 and 311504/2017-5; and Fundação de Amparo à Pesquisa do Espírito Santo -- Brazil (FAPES) -- grant 84412844/2018.}
}

\author{
  \IEEEauthorblockN{
    Lucas C. Possatti\IEEEauthorrefmark{1}\IEEEauthorrefmark{3}, Rânik Guidolini\IEEEauthorrefmark{1}, Vinicius B. Cardoso\IEEEauthorrefmark{1}, Rodrigo F. Berriel\IEEEauthorrefmark{1}, \\
    Thiago M. Paixão\IEEEauthorrefmark{1}\IEEEauthorrefmark{2}, Claudine Badue\IEEEauthorrefmark{1}, Alberto F. De Souza\textit{, Senior Member, IEEE}\IEEEauthorrefmark{1} and Thiago Oliveira-Santos\IEEEauthorrefmark{1}
  }
%  \IEEEauthorblockA{
%    \textit{Departamento de Informática} \\
%    \textit{Universidade Federal do Espírito Santo}\\
%    Vitória, ES, Brazil \\
%  }
	\IEEEauthorblockA{\IEEEauthorrefmark{1}Universidade Federal do Esp\'irito Santo, Brazil}
	\IEEEauthorblockA{\IEEEauthorrefmark{2}Instituto Federal do Esp\'irito Santo, Brazil}
	\IEEEauthorblockA{\IEEEauthorrefmark{3}Email: lucas.possatti@lcad.inf.ufes.br}
}

% \author{\IEEEauthorblockN{Lucas Caetano Possatti}
% \IEEEauthorblockA{\textit{Departamento de Informática} \\
% \textit{Universidade Federal do Espírito Santo}\\
% Vitória, ES, Brazil \\
% email address}
% \and
% \IEEEauthorblockN{2\textsuperscript{nd} Given Name Surname}
% \IEEEauthorblockA{\textit{dept. name of organization (of Aff.)} \\
% \textit{name of organization (of Aff.)}\\
% City, Country \\
% email address}
% \and
% \IEEEauthorblockN{3\textsuperscript{rd} Given Name Surname}
% \IEEEauthorblockA{\textit{dept. name of organization (of Aff.)} \\
% \textit{name of organization (of Aff.)}\\
% City, Country \\
% email address}
% \and
% \IEEEauthorblockN{4\textsuperscript{th} Given Name Surname}
% \IEEEauthorblockA{\textit{dept. name of organization (of Aff.)} \\
% \textit{name of organization (of Aff.)}\\
% City, Country \\
% email address}
% \and
% \IEEEauthorblockN{5\textsuperscript{th} Given Name Surname}
% \IEEEauthorblockA{\textit{dept. name of organization (of Aff.)} \\
% \textit{name of organization (of Aff.)}\\
% City, Country \\
% email address}
% \and
% \IEEEauthorblockN{6\textsuperscript{th} Given Name Surname}
% \IEEEauthorblockA{\textit{dept. name of organization (of Aff.)} \\
% \textit{name of organization (of Aff.)}\\
% City, Country \\
% email address}
% }

\maketitle

\begin{abstract}
Autonomous terrestrial vehicles must be capable of perceiving traffic lights and recognizing their current states to share the streets with human drivers. Most of the time, human drivers can easily identify the relevant traffic lights. To deal with this issue, a common solution for autonomous cars is to integrate recognition with prior maps. However, additional solution is required for the detection and recognition of the traffic light. Deep learning techniques have showed great performance and power of generalization including traffic related problems. Motivated by the advances in deep learning, some recent works leveraged some state-of-the-art deep detectors to locate (and further recognize) traffic lights from 2D camera images. However, none of them combine the power of the deep learning-based detectors with prior maps to recognize the state of the relevant traffic lights. Based on that, this work proposes to integrate the power of deep learning-based detection with the prior maps used by our car platform IARA (acronym for Intelligent Autonomous Robotic Automobile) to recognize the relevant traffic lights of predefined routes. The process is divided in two phases: an offline phase for map construction and traffic lights annotation; and an online phase for traffic light recognition and identification of the relevant ones. The proposed system was evaluated on five test cases (routes) in the city of Vitória, each case being composed of a video sequence and a prior map with the relevant traffic lights for the route. Results showed that the proposed technique is able to correctly identify the relevant traffic light along the trajectory.
\end{abstract}

\IEEEpubid{\begin{minipage}[t][10cm][t]{\textwidth}\ \\[10pt] \centering
  % {\normalsize\textcopyright 2019 IEEE. Personal use of this material is permitted.  Permission from IEEE must be obtained for all other uses, in any current or future media, including reprinting/republishing this material for advertising or promotional purposes, creating new collective works, for resale or redistribution to servers or lists, or reuse of any copyrighted component of this work in other works.}
  {\footnotesize\textcopyright 2019 IEEE. Personal use of this material is permitted.  Permission from IEEE must be obtained for all other uses, in any current or future media, including reprinting/republishing this material for advertising or promotional purposes, creating new collective works, for resale or redistribution to servers or lists, or reuse of any copyrighted component of this work in other works.}
\end{minipage}}

\section{Introduction}

% Alternativa: Fazer um lero-lero de veículos autônomos, e falar que dentre várias partes fundamentais está o "Traffic light recognition".

% Possível lero-lero:
%  - TLs poderiam se comunicar por radio, mas renovar a infraestrutura do local é caro.
%  - Segurança é muitíssimo importante.
%  - Veículos autonomos tem potencial para serem mais seguros que humanos dirigindo.
%  - TLs são estáticos, então pode anotar as posições deles.

Autonomous driving is an essential topic of research in the development of intelligent transportation systems.
Briefly, the great ambition with autonomous vehicles is fully replacing the human driver by a computer system without compromising and eventually improving safety and efficiency. To this end, the human ability of ``seeing'' the environment (e.g., the road, pedestrians,
signs, and other vehicles) and behaving accordingly should be carefully reproduced by the computer system.
In particular, autonomous terrestrial vehicles must be capable of perceiving traffic lights and
recognizing their current states (red, yellow, green).

%overcome, such as: detection in adverse conditions (e.g., rain, snow), early detection (i.e., detecting traffic lights at greater distances),
%and detection in different illuminations settings (including night images).

In the autonomous driving literature, the general problem of identifying traffic lights and their states is known as Traffic Light Recognition (TLR). Although TLR is widely addressed in the literature, there are still some challenges to be faced. Most methods \cite{de2009traffic,de2009real,siogkas2012traffic,philipsen2015traffic} focus on locating and/or recognizing all traffic lights in a scene and do not attribute any special meaning to the traffic lights that are relevant for the given context, i.e., the traffic lights in the vehicle's route and that the driver should obey. Other challenges include recognition in adverse conditions (e.g., rain, snow), early recognition (detecting traffic lights at greater distances), and recognition in different illumination settings (including night images). Despite their importance, these other challenges are out of the scope of this work.

%In a given moment, there can be more the one relevant traffic light, but their states are necessarily concordant. 
%n other words, there is no ambiguous state in any instant for the planned route.

%Choosing what are the relevant traffic lights in a scene is a difficult task that involves a lot of subjectivity.
% It depends on the traffic light position in the image, which lane the car is traveling, and what is the planned route.
%This problem is solved by systems in the literature that integrate prior maps of traffic lights' position and properties.
%As \cite{jensen2016vision} indicates, these systems are among the best performing solutions for the problem of traffic light recognition. 
%With prior maps, an autonomous vehicle that knows its location in the world can know in advance when it is coming close to a traffic light, and where they will appear on camera. This makes it easier to detect traffic lights by filtering out false positives or by detecting traffic lights only within a ROI (region of interest).

%This is a limiting factor of the methods relying solely on image data \lcp{refs} to recognize
%relevant traffic lights. 
%This makes it easier to detect traffic lights by filtering out false positives or by detecting traffic lights only within a ROI (region of interest).

Most of the time, human drivers can easily identify the relevant traffic lights. Nevertheless, there are not always precise rules (e.g., an algorithm) that allow the differentiation of these traffic lights from the others in a scene. To deal with this issue, a common solution is to integrate recognition with prior maps which record position, direction, and other properties of the traffic lights \cite{fairfield2011traffic,barnes2015exploiting,jensen2016vision}. With prior maps, an autonomous vehicle can be early aware of the presence of traffic lights on its vicinity, and can also fuse the map and real-time sensors' data (e.g., camera image) for robust location and recognition of the relevant traffic lights in a scene.

% >>>>> gap <<<<<<
Deep learning techniques have showed great performance and power of generalization in many areas and types of problems such as classification \cite{berriel2017grsl,Berriel2017cag} and detection \cite{GuidoliniIJCNN2018}. General purpose object detectors have been well explored for traffic related problems (such as detection of pedestrians, traffic signs, etc), and YOLO \cite{redmon2016you} and Faster R-CNN \cite{ren2015faster} are two of these state of the art detectors. Motivated by the advances in deep learning, some recent works \cite{weber2016deeptlr,behrendt2017deep,bach2018deep,pon2018hierarchical} leveraged some state-of-the-art neural detectors to locate (and further recognize) traffic lights from 2D camera images. Other works, such as \cite{john2014traffic}, combined prior maps with deep learning classification. However, to the best of our knowledge, none of them combine the power of the deep learning-based detectors with prior maps to recognize the state of the relevant traffic lights. Although they are very powerful object detectors, they cannot identify the relevant traffic light for the vehicle, which is a fundamental task for autonomous driving. 

\begin{figure}[t!]
	% Essa imagem é da entrada 1.
	\centering
	\includegraphics[width=\columnwidth]{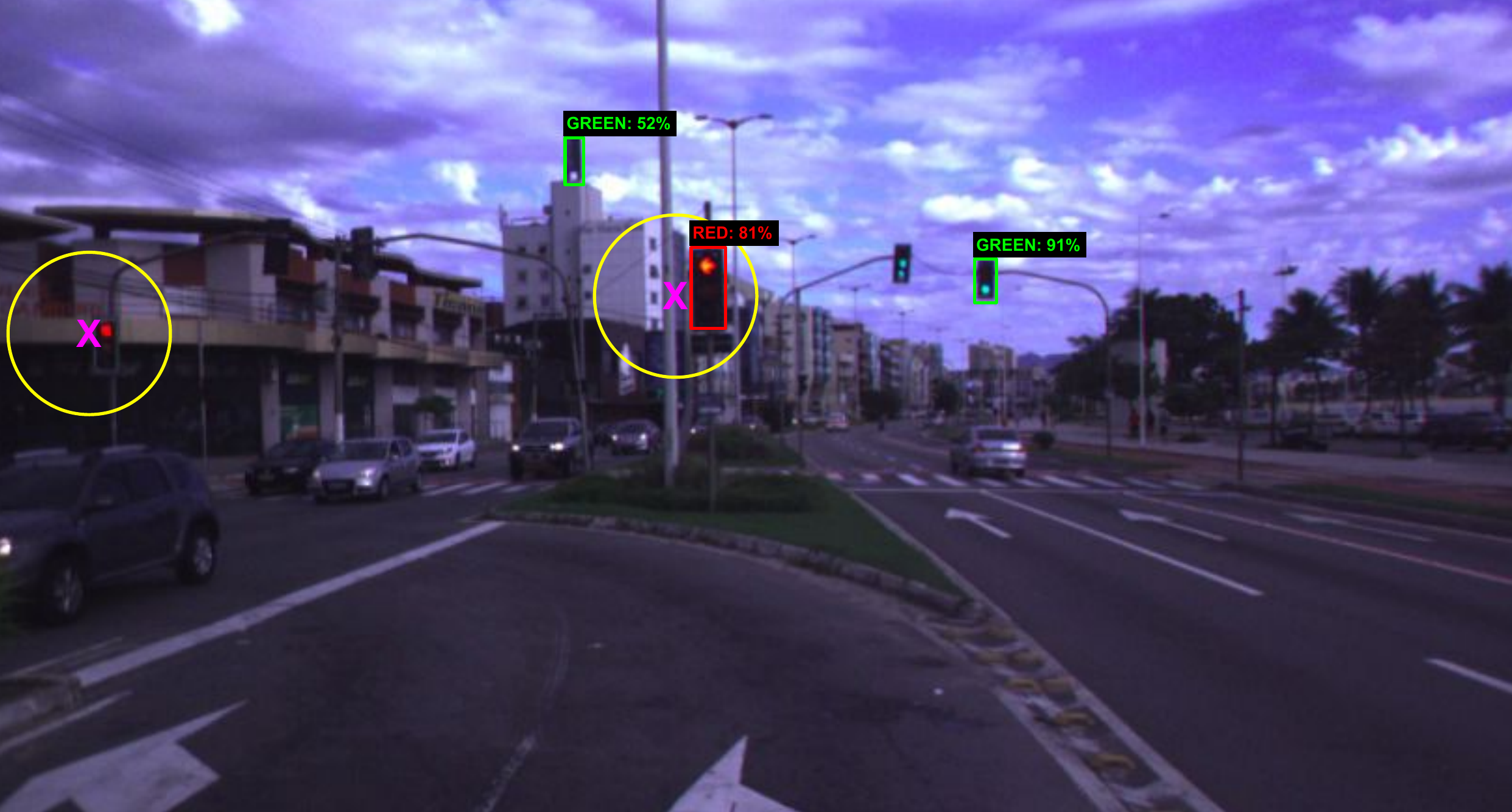}
	\caption{Example of the proposed system running. The traffic lights delimited by bounding boxes (BBs) are located and recognized by a deep detector in a online phase. The text above BBs indicates the traffic lights' state followed by the confidence score of the detection. The pink crosses represent the projection of the mapped (during an offline phase) relevant traffic lights onto the 2D camera space, whereas the yellow circles define a threshold limit for these traffic lights. BBs whose center lies outside the yellow circles are filtered out. The BB closest to any of the red dots will dictate the traffic light (group) state.}
	\label{fig:system-running}
\end{figure}

% >>>>> proposal <<<<<
Based on this need, this work proposes to integrate the power of deep learning-based detection with the prior maps used by our car platform IARA (acronym for Intelligent Autonomous Robotic Automobile \cite{baduesurvey}) to recognize the relevant traffic lights of predefined routes. The process is divided in two phases: an offline phase for map construction and traffic lights annotation; and an online phase for traffic light recognition and identification of the relevant ones.
In the offline phase, the prior map is constructed by driving IARA on routes of interest to collect camera and LiDAR data.
Subsequently, a semi-automatic process is applied to these data in order to find 3D coordinates of traffic light candidates, which are further inspected on camera images to filter false positives and identify those that are relevant for the particular route. The relevant traffic lights' positions (in 3D world coordinates) are then stored in the prior map.
In the online phase, as the car approaches traffic lights, their stored positions are projected to the image in order to filter the predictions (location and state) produced by a deep detector. Finally, only the relevant traffic lights are taken into account in the decision-making process. The relevant state is dictated by the closest traffic light to the projected annotations. It is worth noting that this work treats the traffic light state as a two classes problem (red-yellow and green) in order to cope with the lack of yellow samples for a proper training of a three classes detector. See \figurename~\ref{fig:system-running} for an illustration of the system behavior.

% >>>>>> experimental <<<<<
For performance assessment, the proposed system was evaluated on five test cases (routes) in the city of Vitória, each case being composed of a video sequence and a prior map with the relevant traffic lights for the route. The performance of the system was measured in two ways: (i) how good were the predicted bounding-boxes, i.e., the performance of the detector itself in terms of the mean Average Precision, and (ii) how accurate were the states predicted by the full system, i.e., the ability of the system to correctly assign the state of the relevant traffic lights during the car progress.
  % The proposed approach achieved an accuracy of \lcp{XX.XX\%} on average.

The rest of this paper is organized as follows: related works are discussed in Section \ref{related-work}; the proposed method is described in Section \ref{proposed-method}; the experimental methodology is presented in Section \ref{experimental-methodology}; results are shown and discussed in Section \ref{results}; and, finally, Section \ref{conclusion} concludes and discusses future work.

\section{Related Work}
\label{related-work}

This section covers the main works addressing TLR based on prior maps or deep learning applied to TLR. For a more comprehensive review that includes other approaches, the reader should refer to the surveys in \cite{diaz2015robust,jensen2016vision}.

Prior maps containing annotations of traffic lights (e.g., position, direction) can be exploited to increase the robustness of TLR. In this context, the work of
Lindner et al. \cite{lindner2004robust} uses maps to recognize traffic lights and they propose a three-stage system: detection (based on handcrafted features), tracking, and state classification. Map information and GPS data can be incorporated into the system to allow triggering the system only near intersections, therefore possibly reducing the amount of false alarm detections. Despite of the use of prior maps, their work does not address explicitly the choice of the relevant traffic lights for the lane.

Fairfield and Urmson \cite{fairfield2011traffic} present an automatic strategy to map traffic lights by fusing the precise location of the car with
image information, and then estimating traffic lights' 3D positions using least squares triangulation. Levinson et al. \cite{levinson2011traffic} propose a mapping procedure of traffic lights using tracking, back-projection, and triangulation. Traffic lights' states are computed in a probabilistic approach taking into consideration the previously constructed map.
In addition to prior maps, Frank et al. \cite{franke2013making} locate
relevant traffic lights by matching image-based features of intersections. Such features are extracted offline from manually labeled regions around the relevant traffic lights while a neural network is responsible for the final state classification.

John et al. \cite{john2014traffic} combine prior maps and GPS to limit the region of interest (search area) in which traffic lights are expected to appear. A convolutional neural network (CNN) is used to detect and classify the traffic light candidates within the search area. Jang et al. \cite{jang2017traffic} propose a recognition system that explores the prior maps at every stage. Besides reducing the search area (as in \cite{john2014traffic}), prior maps include the type of traffic light face that helps state classification. They additionally propose a slope compensation method to treat recognition in inclined roads.
The interest in deep learning methods for TLR has been observed in the recent years. Weber et al.\cite{weber2016deeptlr} propose a network called DeepTLR for detection and classification of traffic lights. Behrendt et al. \cite{behrendt2017deep} modified YOLO \cite{redmon2016you} to detect traffic light candidates. A custom CNN is leveraged for state classification. A more detailed study on traffic light detection using YOLO can be found in \cite{jensen2017evaluating}. Pon et al. \cite{pon2018hierarchical} detect simultaneously traffic lights and traffic signs with a modified version of the Faster R-CNN \cite{ren2015faster}.

\begin{figure*}[ht]
    \centering
    \includegraphics[width=0.95\textwidth]{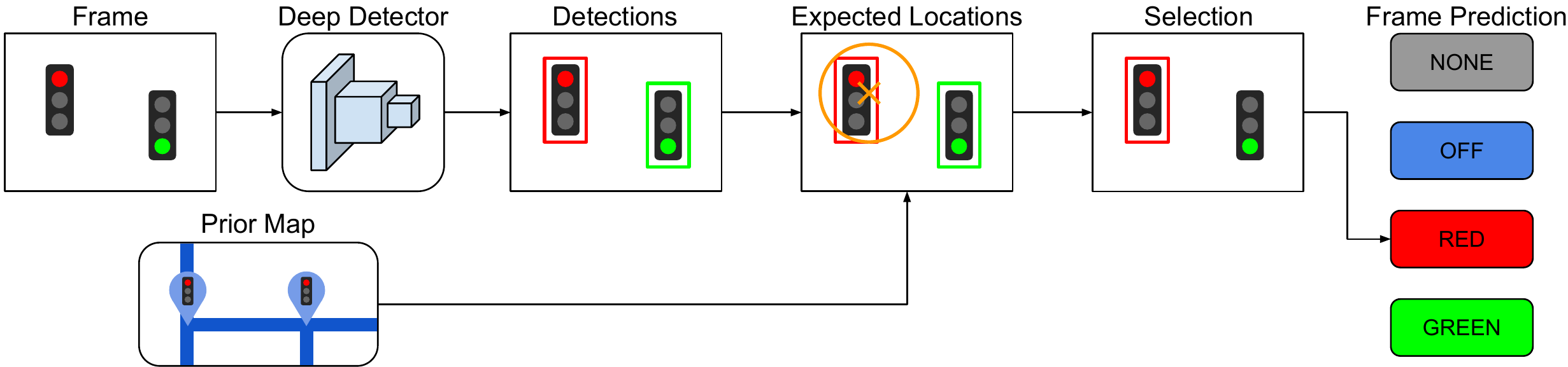}
    \caption{Overall flow of the system. First, a camera frame is fed to the deep learning detection model, which proposes bounding boxes with TLs' state. Subsequently, TL's world positions from the prior maps are projected into the image (orange cross) using the current localization pose of the vehicle. The orange circumference represents a threshold that accounts for imprecisions in localization. Finally, from BBs that have center within the threshold, the closest to any of the projected TL's positions is selected and used as final state prediction for that frame.}
    \label{fig:system_inference}
\end{figure*}

Despite the fact that TLR can benefit from modern deep learning techniques, additional information should be used to allow detection/classification of the relevant traffic lights. Therefore, this work proposes to combine a state-of-the-art deep detector with precise localization (e.g., by exploiting LiDAR data) provided in prior maps for a real-world TLR application in autonomous vehicles. 
%
% with a deep learning network to detect traffic lights and classify their state, and prior maps to select only relevant traffic lights. We also explore using a LiDAR sensor to create prior maps.

%Their best model achieves 95.6\% of precision and 91.4\% of recall with 25\% of IoU on images of $1280\times960$ pixels from a local dataset. The model takes 80 ms to process an image of $1280\times960$ pixels and 28.8 ms for $640\times480$ pixels.

% that are further classified and tracked over frames. For images of resolution of $1280\times720$ pixels, they make three crops that cover the upper part of the image, and feed each crop to YOLO with an input resolution of $448\times448$ pixels. The detector alone runs at 10 FPS.
 % They also contribute with a new dataset of traffic lights with more than 5000 $1280\times720$ pixels color images captured in San Francisco, California.
 % Affs, os resultados de behrendt2017deep são gráficos de precision-recall, assim fica difícil reportar um número aqui.

% % (A)
% Following the track laid by the previously mentioned works we implement a similar approach on IARA, our autonomous vehicle. We use a deep learning network to detect traffic lights in the camera frames, and choose the relevant ones based on prior map information and an accurate localization system.

% (B)

\section{Traffic Light Recognition for Autonomous Vehicles}
\label{proposed-method}

This work proposes a method for recognition of relevant traffic lights (TLs) and their state using a deep learning detection model and prior maps of an autonomous vehicle. The process is performed in two steps: offline map construction and traffic light annotation for the relevant lane, and an online detection and recognition of the relevant traffic light state. \figurename~\ref{fig:system_inference} displays an overall picture of the online phase of the method. The detection model takes camera frames as input and proposes bounding boxes (BBs) with the respective TL state. Subsequently, information from prior maps and the localization of the vehicle is used to project the TL world position to the current frame and then select only one BB to make the final state prediction for that frame. Additionally, a new method for creating prior maps that uses both the detection model and the LiDAR sensor is proposed. The proposed approach is implemented in the autonomous car of our laboratory, IARA.

% The next subsection describe: the IARA platform, autonomous car where the proposed approach was implemented; the traffic lights detection and recognition, the deep learning detector used to find the traffic lights in the frame and recognize their state; the generation of prior maps, used to annotate and store the relevant traffic light world position for a given route; and finally, the general system operation, that describes how this steps are connected.
The next subsections describe: the IARA platform, the autonomous car on which the proposed approach was implemented; the traffic lights detection and recognition procedures, which includes the deep learning detector used to find the traffic lights in the frame and recognize their state; the generation of prior maps used to annotate and store the relevant traffic light world position for a given route; and, finally, the general system operation, that describes how these steps are connected.

%The basic assumption of the proposed approach is that the definition of the relevant traffic lights is tied to the specific route taken by the car. To construct the prior maps, a two-stage procedure is employed. First, our car platform IARA is driven (acronym to Intelligent Autonomous Robotic Automobile \cite{baduesurvey}) on routes of interest to collect camera data, as well as 3D LiDAR data that provides accurate localization of obstacles (including traffic lights) along the way. In a second stage, a human annotator identifies the relevant traffic lights in the camera images. These annotations are then merged with the LiDAR data (projected onto 2D camera space) resulting on a map with the traffic lights locations. At test time, the previously annotated positions of the traffic lights (i.e., the prior maps) are used to filter the predictions (i.e., location and state) in the 2D image produced by a deep detector, allowing only the relevant traffic lights to be taken into account in the decision-making process (\Cref{fig:system-running}).

% OVERVIEW: We describe the process we use for creating prior maps, and how this information is used to aid in the traffic light recognition task.

\subsection{Intelligent Autonomous Robotic Automobile (IARA)}
\label{IARA}

IARA is the autonomous car built in our laboratory. It is an adapted Ford Escape Hybrid that features a variety of sensors such as: odometer, LiDAR, IMU, RTK-GPS, and stereo cameras.
The LiDAR is a Velodyne's HDL-32E and the stereo camera is a Point Grey's Bumblebee XB3 stereo camera. Both are mounted on top of the car, and the camera is mounted front-facing. The HDL-32E LiDAR uses 32 lasers spread at a 40° VFOV (+10° to -30°) to scan 360° horizontally. The Bumblebee XB3 camera has three 1.3MP CCD sensors with % 3.8mm of focal length and
66° of HFOV, and captures colored images of $1280\times960$ pixels at 16 FPS.
% FIXME: Talvez eu não precise entrar em tanto detalhe assim dos sensores. Mas eu vi muito paper detalhando bastante as cameras, por exemplo.

IARA’s software is based on the Carnegie Mellon Robot Navigation Toolkit (CARMEN), which is an open source collection of software for robot control \cite{montemerlo2003perspectives}.
Our laboratory, LCAD, maintains its own fork of CARMEN at \url{https://github.com/LCAD-UFES/carmen_lcad}.

% TODO: Pedir ao Mutz pra revisar esse parágrafo se ele puder.
To localize the car accurately in the world, GPS coordinates are often not enough. For this reason IARA uses a localization system \cite{de2016light} based on Particle Filter localization.
The system is initialized with a pose from GPS and orientation from IMU, it then enters a cycle of two phases: prediction and correction. In the first phase, the system predicts car poses using odometry data. In the second phase, the method corrects these poses by matching 2D local occupancy grid maps with a global one that is generated offline using the technique presented in \cite{Mutz:2016:LMC:2873073.2873265}.
In order to create the 2D grid maps, the localization system transforms 3D point clouds from LiDAR into 2D grid maps.
And, to create the global map, IARA applies a SLAM-based system, which uses data from odometer, GPS, IMU, and LiDAR. 
Experiments showed that IARA's localization system operates within 0.28 meters of longitudinal error and 0.14 meters of lateral error \cite{de2016light}.

% \subsection{Localization}

% For every possible track run by IARA, we want to use 3D coordinates for each traffic light it is going to find on its way. Therefore, % or consequently
% we need IARA to localize itself in the world with good amount of precision.
% To localize the car's position in the world, GPS coordinates alone are not sufficient, since they have some imprecision that makes them unsuitable for such task. In order to properly localize itself, IARA uses a combination of GPS coordinates, odometry, and LiDAR. After recording a driving log (a recording of sensors for a given run), we can use all this data to build a, somewhat precise, cost map. The cost map is a grid where each cell contains a probability of being occupied or not, which is obtained from LiDAR data. After building the map, we know for each cell its coordinates in the world with a precision of \lcp{====}.
% On subsequent runs on this map, IARA can use all the sensor data mentioned and \lcp{some weird algorithm I don't remember the name} to find its precise position in world coordinates.
% % Using the cost map, combined with LiDAR information, GPS coordinates, and odometry, IARA can localize itself properly in the world.

\subsection{Traffic Lights Detection and Recognition}

A deep learning neural network model (YOLO \cite{redmon2016you}) is used for detecting traffic lights and classifying their state. The YOLOv3 (the third version of the YOLO) was chosen because it is one of the state-of-the-art detectors and it can achieve good performances with high frame rate.
In this work, only two classes of objects are considered: red-yellow and green traffic light. The choice of mixing red and yellow in one class was to overcome the lack of yellow samples to proper train a three classes detector. Most of the traffic light databases provide very few samples of yellow traffic lights. For the purposes of our application, joining these classes is acceptable since it is better to have IARA stopping before a yellow traffic light than proceeding.

% TRAINING: The input for a deep learning detection model is usually: a colored image; and a ground truth list of bounding boxes with the object's class.
Deep learning detection models usually take a colored image as input for inference. The network forwards the image through the convolutional layers of a standard convolutional neural network (e.g., ResNet, or Darknet) for feature extraction. Further layers process the output of this feature extraction
and generate a list of bounding boxes (BBs) with the object's class probabilities. These are the final outputs of the network.

% TODO: Possívelmente falar mais sobre CNNs e camadas convolucionais, se estiver faltando escrever mais. john2014traffic fez isso.

\subsection{Prior Maps Generation}

% When driving autonomously, IARA will always follow a predetermined path from a Road Definition Data File (RDDF), which is generated offline.
% For a given RDDF, we annotate the 3D location of each traffic light relevant for the path it will follow. Hence, whenever IARA is following that RDDF, it will know in advance where the traffic lights are supposed to be, and where they should appear on camera.
% % , so that we can find their bounding boxes and consequently their state.

% % TODO: Falar sobre o agrupamento de TLs.

% In order to build prior maps, we drive IARA manually along the route we want it to repeat in the future, recording all sensor data (we call it a "log"). The sensor data is later used offline to build the RDDF, and the global occupancy grid map described in \ref{IARA}. Then, yet offline, we replay the log, while doing traffic light detection. We project LiDAR rays to camera coordinates, select only those that "hit" inside a bounding box, and accumulate them throughout the log's execution.
% When eight frames pass without a single detection, the accumulated point cloud is clustered using the Density-Based Spatial Clustering of Applications with Noise (DBSCAN) algorithm \cite{ester1996density}, yielding centroids that represent TL positions. The accumulated point cloud is then reset, and the process repeats through the log's duration.
%  Also, we have to manually attribute TL positions to the RDDFs where those TLs are relevant.
% % For each 

The standard mode of operation of IARA is to follow a predetermined path from a Road Definition Data File (RDDF) when driving autonomously. The RDDF is a trajectory that was performed by a human operator driving the car and it stores information like speed, position in the lane, etc. As described in Section \ref{IARA}, IARA also needs a global occupancy grid map to properly localize itself in the world. To generate this information, IARA has to be driven manually along the route of interest recording all sensor data (referred here as "log"). The sensor data is later used, in a offline process, to build the RDDF and global occupancy grid map.

Given that the recorded log is already necessary for the standard IARA's operation, it can also be used to create the prior map of relevant traffic lights for the respective RDDF.
Making use of prior maps is necessary because until the present moment there is no clear algorithm or machine learning method that can robustly identify which traffic lights are relevant using only image data.
The first step is to identify the traffic lights' positions in the world. For that, the log is played offline, while the traffic light detector locates the traffic lights in the image frames. Additionally, the LiDAR points are projected to camera coordinates, and those that "hit" inside any traffic light bounding box are accumulated in a buffer. The points are projected using the same process as presented in \cite{GuidoliniIJCNN2018}. When eight frames have passed without a single detection, the accumulated point cloud is clustered using the Density-Based Spatial Clustering of Applications with Noise (DBSCAN) algorithm \cite{ester1996density}. The world position of the cluster centroids are used as traffic lights' positions. The accumulated point cloud is then reset, and the process repeats throughout the duration of the log.
Finally, the centroids are manually filtered to discard false positives and traffic lights that are not relevant for the RDDF of interest.
Moreover, traffic lights that share the same control semantics can be grouped together, thus their redundancy can come to use, i.e., when the state of one of them can not be determined, the state of the other is used instead. Traffic lights that are relevant for a particular route are grouped considering a maximum distance threshold (of 20 meters) from each other. One important feature is that the annotated traffic lights can be transferred from one RDDF to the other, for example, when two RDDFs follow the same path (maybe in different lanes). Additionally, the traffic lights that were discarded for not being relevant for one RDDF could be reused for the prior map of another RDDF where they are relevant.

% sadsadsadsadsadsadasegf we w f gr er w gwt gw hwh  tr hwr
% sdfsdafbsdkafbs dkafbsdjkabfs djkafsdj kabngfksjd
% sdfsdafbsdkafbs dkafbsdjkabfs djkafsdj kabngfksjd
% sdfsdafbsdkafbs dkafbsdjkabfs djkafsdj kabngfksjd
% ukgbdkfg sdf sda fsda sda fsd fsdfa adg dfg fda
This clustering strategy has to cope with some difficulties:  
(i) TLs are small objects and LiDAR rays are sparse, thus, in some cases, only few rays will hit a specific TL or perhaps none at all;
(ii) small imprecisions in the localization system may cause disturbances in the accumulated point cloud;
(iii) if the BB is a little larger than the TL object, then some LiDAR rays may hit inside the BB, but they actually miss the real TL object, hitting something on the background instead. Thus, it is better to train the detector for the offline phase with tight BB annotations, so that BB proposals will also be tight. All of these difficulties, may cause the clustering algorithm to yield false positives and also some inaccurate TL positions. Therefore, they need to be manually filtered as mentioned. In case of any problem in the final accumulated points, there is always the possibility of choosing a single frame for extracting the traffic light world position. 

\subsection{General System Operation}

When driving autonomously, IARA continuously checks for the relevant traffic lights associated with the RDDF. 
Whenever a group of relevant TLs comes within 100 meters, their 3D world position is projected onto camera coordinates. Additionally, the detector is triggered for the current frame. Each annotated TL position is surrounded by a sphere of 1.5 meter of radius that is also projected to serve as a threshold for the localization error. The euclidean distance, in camera's coordinates, between each BB's center and each projected TL is calculated. Any BB that has its center outside of all projected spheres are discarded right away. From the remaining ones, that with the closest center to any of the projected TLs is selected, and its status serves as the final prediction for that frame. The final state output of the system is either one of the four: none, when there is no traffic light; off, when there is a traffic light, but the state is not recognized, red, when it is red or yellow; and, green, when it is green. In cases in which all BBs are discarded, the final state is set to "off". Before a red, yellow or off traffic light, IARA will reduce its speed until a complete stop if necessary, otherwise, before a green traffic light, it will continue its trajectory.

\section{Experimental Methodology}
\label{experimental-methodology}

The experiments for this work aim mainly to evaluate full TLR system on five driving logs. In addition to the traditional metrics, we also analyzed how early the traffic lights are correctly detected. As preliminary study, we also included the performance evaluation of the deployed detector (an YOLOv3 model). The rest of this section describes the datasets used in the experiments, the driving logs (used to reproduce the IARA's sensors offline), the performance metrics, and the experiments themselves.
%
%To evaluate the proposed system, we measured how much the predictions diverged from ground truth, and examine how early it can predict the traffic light state correctly.

\subsection{Datasets for Traffic Light Detection}

The training and evaluation of the YOLOv3 detector leveraged two public available datasets: DriveU Traffic Light Dataset (DTLD) \cite{fregin2018driveu} and LISA Traffic Light Dataset (LISA-TLD) \cite{jensen2016vision}.

The DriveU Traffic Light Dataset (DTLD) comprises traffic image sequences from 11 German cities. Each image ($2048\times1024$ pixels) includes the following annotations about traffic lights: bounding box coordinates, state, relevance status (visually defined by annotators), horizontal or vertical orientation, occlusion, number of lights, type of traffic light (for pedestrians, cyclists, or cars), and other attributes. For our purposes, pedestrians and cyclists traffic lights were removed. From the remaining, only active (i.e., not ``off'') traffic lights which are facing the car and have at least three bulbs were selected. The images are cropped to a $1280\times960$ rectangle aligned to the top and centered on the horizontal axis, and then scaled down to $640\times480$ pixels. The bounding box annotations were changed accordingly to reflect the cropping and scaling. The dataset comes with a training-test split.

The LISA Traffic Light Dataset was developed by the Laboratory for Intelligent and Safe Automobiles (LISA) at University of California.
 % from LISA (Laboratory for Intelligent and Safe Automobiles) at University of California, San Diego.
The dataset was created using a Point Grey's Bumblebee XB3 (the same model in IARA) to record more than 44 minutes of traffic video sequences in San Diego, California, USA. Some video sequences were recorded during the day, and others during the night. The dataset is split into training and testing sets, and has seven TL classes: ``go'', ``go forward'', ``go left'', ``warning'', ``warning left'', ``stop'', and ``stop left''. The classes ``go'', ``go forward'', and ``go left'' were merged as ``green'', and the other classes as ``red'' since we do not address the recognition of yellow traffic lights in separate.

For the full system evaluation, we used a local dataset named (in this work) IARA Traffic Light Dataset (IARA-TLD). It comprises images of traffic scenes recorded in Vitória, Espírito Santo, Brazil with help of the IARA's camera. The dataset has a total of 5002 bounding box annotations distributed on three different classes: ``green'', ``yellow'', and ``red''. The  ``red'' and ``yellow'' classes were also grouped into a single class ``red''. The images were scaled down to $640\times480$ pixels. Table~\ref{tab:bbs-per-class} shows the distribution of the classes (i.e., number of annotations per class -- ``red'' and ``green'') of the annotated traffic lights across the datasets' splits.

\begin{table}[!t]
  % I formatted the table as done in IEEEtran HOW TO guide.
  \renewcommand{\arraystretch}{1.3}
  \caption{Distribution of the annotated traffic lights across the datasets' splits.}
  \label{tab:bbs-per-class}
  \centering
  % \begin{tabular}{|c|c|c|c|c|c|}
  %   \hline
  %   Class & \thead{DTLD\\(train)} & \thead{DTLD\\(test)} & \thead{LISA-TLD\\(train)} & \thead{LISA-TLD\\(test)} & IARA-TLD \\ \hline\hline
  %   Red   & 20936                 & 9147                 & 23129                     & 10303                    & 2079     \\ \hline
  %   Green & 33540                 & 14403                & 14681                     & 7717                     & 2923     \\ \hline
  % \end{tabular}
  \begin{tabular}{@{}lccccc@{}}
    \toprule
    % Class & \thead{DTLD\\(train)} & \thead{DTLD\\(test)} & \thead{LISA-TLD\\(train)} & \thead{LISA-TLD\\(test)} & IARA-TLD \\ \midrule
    Class & \mergeh{c}{c}{DTLD\\(train)} & \mergeh{c}{c}{DTLD\\(test)} & \mergeh{c}{c}{LISA-TLD\\(train)} & \mergeh{c}{c}{LISA-TLD\\(test)} & IARA-TLD \\
    \midrule
    Red   & 20936                 & 9147                 & 23129                     & 10303                    & 2079     \\
    Green & 33540                 & 14403                & 14681                     & 7717                     & 2923     \\
    \bottomrule
  \end{tabular}
\end{table}

\subsection{Driving Logs}

Driving logs are files containing detailed sensory information recorded during the car's trips, and serve to reproduce the IARA's sensors offline. 
%A human drives the car, and IARA's software records data from all sensors, so that they can be later replayed offline.
These logs can be revisited (playback operation) as many times as needed, which allows to test offline modifications on the IARA's software, and evaluate the new car's behavior. A total of 10 logs were recorded on Dante Michelini avenue (Vitória, Espírito Santo, Brazil), from which five were used to construct the prior maps, and the other five for full system evaluation. This avenue was chosen because it contains challenging situations, i.e., several traffic lights where some of them are potentially discordant. There are four bifurcations, each one with three lanes that go straight, and two other lanes (the two leftmost) turning left. In case of turning left, the driver should take one of the two leftmost lanes and obey the corresponding traffic lights. Otherwise, the driver can keep on the three rightmost lanes ruled by the other set of traffic lights. Despite of the current lane, all traffic lights are visible to the camera most of the time.  Fig.~\ref{fig:system-running} depicts this ambiguous situation.

\begin{figure}[t!]
  \centering
  \includegraphics[width=\columnwidth]{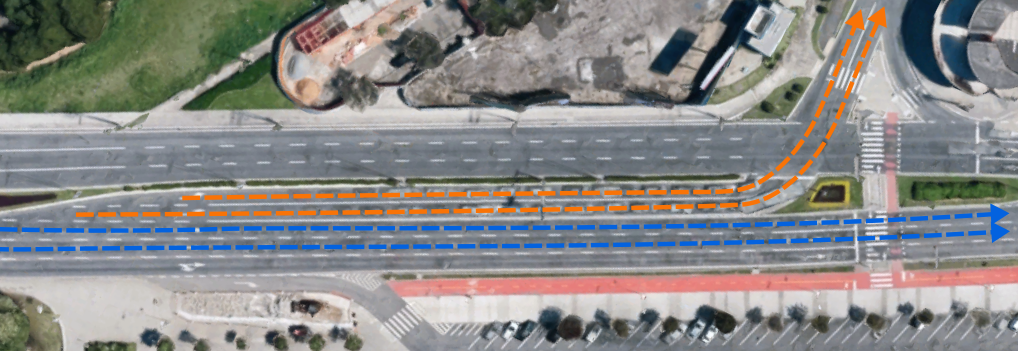}
  \caption{Satellite photo of the first stretch. Each path drawn represents a log. The orange ones are local to that stretch, but the blue ones go through the entire avenue.}
  \label{fig:entrada1}
\end{figure}

%We recorded a total of ten driving logs. To refer to each log individually throughout this work, we attribute names to each of them.

The logs (a total of 10) are associated with stretches and lanes in the avenue. The first stretch is showed in Fig.~\ref{fig:entrada1}. The lanes represented by the blue arrows follow all the avenue. The leftmost (from the car view) is identified by RL, and the other (the middle lane) by RM. The orange arrows are restricted to this stretch, being the leftmost identified by LL-1, and the rightmost by LR-1, where 1 is related to the first stretch.
The rest of the LL/LR lanes are numbered according to their corresponding stretch.

% represent the 

%The three rightmost lanes follow the avenue from beginning to end, RL is the %one on the left, and RM is the one on the middle.
%Each of the difficult stretches has two lanes on the leftmost side that lead to turning left. From these two, LL is the left lane and LR is the right lane, and a number is used to indicate the stretch (e.g, LL-4).
%  % E.g., LR-1 is the log recorded on the right lane from those two that lead to turning left on the first stretch.
%
%It's important to note that none of the logs were recorded in any of the roads present in IARA-TLD.

% We recorded a total of ten driving logs.
% Some of these logs were used for generating prior maps of relevant traffic lights and calibrating the system, while the others were used for testing it.
% On the three rightmost lanes (those that follow straight), we recorded a driving log in the left lane (RL) and middle lane (RM) that follow from the beginning of the avenue until the end.
% The left lane was used for testing, and the middle lane was used to build the global occupancy grid-map of the region (\ref{IARA}) and the prior traffic light maps.
% In each of the four difficult stretches, we recorded logs on each of the two leftmost lanes (those leading to turn left). The logs on the right lane were used for testing, and those on the left lane were used for building prior maps.

% \lcp{TODO. Colocar um mapa com as rotas de cada log. Talvez com zoom em cada uma das entradas.}

% \lcp{TODO. Colocar imagens da camera mostrando as quatro entradas.}

\subsection{Metrics}

The detection performance was measured in terms of precision, recall, and mean average precision (mAP).
  % Precision and recall are defined \eqref{eq:precision} and \eqref{eq:recall} (TP, FP, and FN mean, respectively, true positives, false positives, and false negatives).
These metrics were calculated using an IoU (Intersection over Union) threshold of 0.5, and for a confidence threshold ($\tau$) of 0.2 and 0.5.

% \begin{equation}
% \label{eq:precision}
% Precision = \frac{TP}{TP + FP},
% \end{equation}

% \begin{equation}
% \label{eq:recall}
% Recall = \frac{TP}{TP + FN}
% \end{equation}

The mAP calculation follows the definition for the Pascal VOC 2007 competition \cite{everingham2010pascal}. Basically, mAP is defined as the mean of the Average Precision (AP) for each class, which, in turn, is the value corresponding to the area under a precision-recall curve.
Pascal VOC uses an IoU threshold of 0.5 in the APs' calculation.

For the full system evaluation, the state of the relevant traffic lights was predicted for all the frames of each test log. We reported the confusion matrix resulting from comparing the predictions with the ground truth annotations. Additional information was recorded (for each stretch/lane) in order to verify how early the system correctly perceived the traffic lights: the time the system took to produce the first correct prediction since the car entered the 100-meters range from the next set of traffic lights; the distance the car is when this occurred.

\subsection{Experiments}

Two different experiments are performed in this paper. First, a deep learning model (YOLOv3) for detecting traffic lights and their state was trained and evaluated.
 %Secondly, we annotated 3D coordinates for all the relevant traffic lights of each log used for testing, and replayed the logs to measure the system performance.
Second, the proposed system was tested using the driving logs on avenue Dante Michelini.

\subsubsection{Training the Detection Model}

Currently there are easily available pre-trained models that can detect traffic lights, such as those trained on the COCO dataset \cite{lin2014microsoft}. But, usually, these do not identify the traffic light's state, so we trained our own model.
The YOLOv3 \cite{yolov3} was trained for detecting only traffic lights and their state. The model was trained using the training sets of the DTLD and LISA-TLD. The model was validated on the LISA-TLD's test set using the Pascal mAP metric.

YOLO is a deep learning architecture for object detection, first implemented on the Darknet framework. The third version of this architecture, YOLOv3, uses Darknet-53 (a CNN model with 53 convolutional layers) as its backbone, and reaches 57.9 mAP ($\textrm{AP}_{50}$) on Microsoft's COCO dataset, using an input resolution of $608\times608$ pixels.
According to YOLOv3's report, it performs around 3.8 times faster than RetinaNet, which is also a good alternative once its best model achieves 61.1 mAP on COCO. RetinaNet's slower inference times make us opt for using YOLOv3, for a more real time solution.

The YOLOv3 was trained for 15000 batches, with 64 images per batch, and a constant learning rate of $10^{-4}$.
  % After \lcp{XX} batches, we reduced the learning rate from \lcp{XX} to \lcp{XX} until the end of the experiment.
The input resolution was $608\times608$ pixels as a compromise between inference time and accuracy.
Many of the default parameters were also kept, such as: image augmentation with changes in hue, saturation, and exposure; batch normalization; and default anchors.
  % \lcp{(recalculating anchors showed similar results, but no gain)}.
Additionally, YOLO was allowed to change the input resolution from $608\times608$ to different resolutions following multiples of 32 starting from 320 (e.g., 320, 352, $\dots$, 608) every 10 batches, as it is done in its original work. This, supposedly, makes the model more robust to different scales.

\subsubsection{Experiments on Logs}

The testing logs on avenue Dante Michelini were used to evaluate the overall system performance. RM was used to build the global occupancy grid-map of the region (Section \ref{IARA}). RM and LL-\{1-4\} were utilized for creating prior traffic light maps for the testing logs, which were RL and LR-\{1-4\}.
The detection model outputs many bounding boxes at different confidence values which range from 0 to 1. A confidence threshold $\tau$ is used to eliminate all bounding boxes that have confidence bellow the threshold. By reducing $\tau$, the detector will output more bounding boxes, causing an increase in recall but lowering precision. Since our method uses prior maps to filter false positives, a lower value for $\tau$ could be profitable. Therefore, two values for $\tau$ are investigated, $\tau \in \{0.5, 0.2\}$.
For each testing log, the final predictions of our system is compared with the ground truth at every frame.

\section{Results} % or "Results and Discussion"
\label{results}

In this section, the evaluation of our experiments are displayed and discussed.
  % First, the evaluation and selection of the deep , then the overall system's
First, the deep detector's evaluation, then, measurements of the system's overall performance.

\subsection{Detection Model}

\begin{table}[!t]
  % I formatted the table as done in IEEEtran HOW TO guide.
  % \renewcommand{\arraystretch}{1.3}
  \caption{Detections Results on DTLD, LISA-TLD and IARA-TLD (\%)}
  \label{tab:detection-evaluation}
  \centering
  \begin{tabular}{@{}clccc@{}}
    \toprule
    & Dataset             & Precision & Recall & $\textrm{AP}_{50}$ \\ \midrule
    \multirow{3}{*}{$\tau = 0.2$}
    & DTLD (test set)     & 80.86 & 91.05 & 85.62 \\
    & LISA-TLD (test set) & 62.81 & 62.49 & 50.59 \\
    & IARA-TLD            & 60.86 & 62.28 & 55.21 \\ \midrule
    \multirow{3}{*}{$\tau = 0.5$}
    & DTLD (test set)     & 88.59 & 86.50 & 85.62 \\
    & LISA-TLD (test set) & 66.45 & 54.82 & 50.59 \\
    & IARA-TLD            & 69.53 & 57.16 & 55.21 \\ \bottomrule
  \end{tabular}
\end{table}

The detection model was evaluated on DTLD, LISA-TLD and IARA-TLD using precision, recall and mAP. A IoU threshold of 0.5 was used, and $\tau \in \{0.2,0.5\}$ for measuring precision and recall, anticipating the values that will be used when evaluating the system. The detector achieves $55.21\%$ of mAP on IARA-TLD and $62.28\%$ of recall at $\tau = 0.2$.
  % Testing on IARA
Results for all the datasets can be seen on Table~\ref{tab:detection-evaluation}.
  % In the table, recall and precision were measured using $\tau \in \{0.2,0.5\}$, anticipating the values we will use when evaluating the system, and IoU threshold of 0.5, the same we use for measuring mAP.
Reducing $\tau$ increases recall, but reduces precision. A higher recall is usually desired when bounding boxes can be further filtered, e.g., using prior maps as in this work. % Pq se não, perde os proposals de vez.

Taking an image of $608\times608$ pixels as input, YOLOv3 takes about 47 milliseconds to make proposals on IARA's video card (Nvidia's Titan Xp). This is roughly 21 Hz, but this is capped by the low frequency of our camera: 16 Hz.
 % 0.047 seconds

\subsection{Entire System}

\begin{figure*}[t!]
  \centering
  \includegraphics[width=\textwidth]{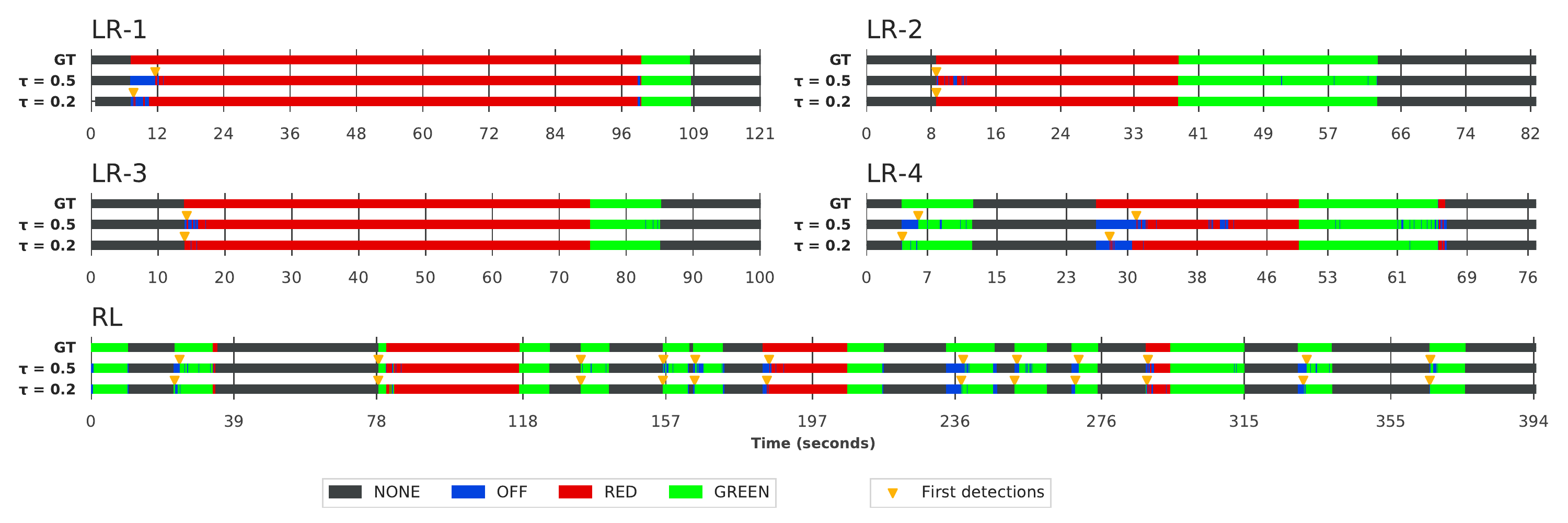}
  \caption{Summary of results on each test log. Each subfigure corresponds to one log, and it shows ground truth and predictions at different confidence thresholds ($\tau$) over time. The earliest detections for each prediction timeline are highlighted.}
  \label{fig:system-results}
\end{figure*}

\begin{table}[!t]
  \caption{First Correct Detections on Logs}
  \label{tab:early-detection}
  \centering
  \begin{tabular}{@{}cccc@{}}
    \toprule
    & Log & \begin{tabular}[c]{@{}c@{}}Delay\\ (seconds)\end{tabular} & \begin{tabular}[c]{@{}c@{}}Distance\\ (meters)\end{tabular} \\ \midrule
    \multirow{5}{*}{$\tau = 0.2$} & LR-1 & 0.56 & 94.47 \\
    & LR-2 & 0.00 & 99.70 \\
    & LR-3 & 0.06 & 99.63 \\
    & LR-4 (I) & 0.07 & 98.42 \\
    & LR-4 (II) & 1.57 & 84.12 \\ \midrule
    \multirow{5}{*}{$\tau = 0.5$} & LR-1 & 4.50 & 60.81 \\
    & LR-2 & 0.00 & 99.78 \\
    & LR-3 & 0.38 & 96.35 \\
    & LR-4 (I) & 1.92 & 78.25 \\
    & LR-4 (II) & 4.63 & 58.75 \\ \bottomrule
  \end{tabular}
\end{table}

To evaluate the system's performance, its predictions are compared with ground truth throughout the frames of each test log.
Fig.~\ref{fig:system-results} shows this comparison in the format of a timeline. As time progresses, the car approaches different traffic lights.
  % For all logs, there are also some frames in each log without traffic lights ("none") within a hundred meters.
For all test logs, there are some frames without traffic lights ("none") within a hundred meters.
  % As it can be seen, most of the predictions are consistent with the ground truth, and the inconsistencies usually occur in the beginning of the detection of a traffic light.
As it can be seen, most of the predictions are consistent with the ground truth, and the inconsistencies usually occur when the car is approaching a traffic light.
The first correct predictions are represented on Fig.~\ref{fig:system-results} as yellow downward triangle markers. Table~\ref{tab:early-detection} also reports the first correct predictions on the shorter logs. On this table, it is possible to see the time the system took to first predict the traffic light state correctly, and the distance remaining until the traffic light is reached. For $\tau = 0.2$, the first detections occur in average with 0.65 seconds of delay and with a distance of around 93 meters  of distance, whereas for $\tau = 0.5$ it occur in average with 1.53 seconds of delay and with around 84 meters of distance.
  % Two different groups of TLs appear on LR-4, so the table reflects that.
 % and how far, the system could first predict the state correctly.
  % At $\tau = 20\%$, correct predictions generally happen a little earlier.

At $\tau = 0.5$, the system predicted the traffic light state correctly most of the time, but its predictions were fickle, due to unstable proposals from the detector. The detector is not confident enough at long distances, which indicates that it cannot handle small traffic lights well. One possible solution to increase stability would be to use higher resolution images. If the detector cannot detect the traffic light from the prior map, an "off" state is emitted. These misses at long distances can be seen in Fig.~\ref{fig:system-results}, for example, at around 10 seconds on LR-2, and at 6 seconds on LR-4. Reducing $\tau$ to 0.2, increases recall, which in turn, leads to improved predictions at long distances as well as reduces the fickle aspect (e.g., LR-4, from 38 to 46 seconds). As a side effect, reducing the threshold also resulted in more false proposals from the detector along the frames, as expected, but these were not a problem since they are normally filtered out by the use of prior maps most of the time.
It is worth noting that a truck passed in front of the traffic light in LR-1 resulting in a occluded traffic light (or off state) that was not annotated, but can be seen in the sequence. Table~\ref{tab:confusion-matrices} reports the confusion matrices for the system at $\tau = 0.2$ so that the reader can have a better idea of the overall prediction performance of the system.

When the car approaches a group of relevant traffic lights, there are two types of error that could occur. The first is predicting the traffic lights are red or off when they are actually green, which represents an inconvenience since it will cause the car to slow down to a stop when it could have proceeded. The second type of error is predicting green traffic lights while facing red ones (or off). This is by far the worst since it could lead the car into the middle of a busy intersection and cause accidents.
  % That is why our system works with the default assumption that, if there are traffic lights nearby, they are off.
  % That is why our system, when it cannot determine the state of a traffic light, assumes it is off.
  % That is why, as a preventive measure, our system assumes traffic lights are off when their state cannot be determined. 
In our experiments, the second type of error was observed, for brief moments and at a long distance, on the second red traffic light of the RL log (see Fig.~\ref{fig:system-results}, between 78 and 118 seconds).

% \begin{table*}[!t]
%   % \renewcommand{\arraystretch}{1.3}
%   \caption{Confusion Matrices for Each Testing Log ($\tau = 0.2$)}
%   \label{tab:confusion-matrices}
%   \centering
% \renewcommand\tabcolsep{5.5pt}
% \begin{tabular}{@{}rcccc|cccc|cccc|cccc|cccc@{}}
% \toprule
%  & \multicolumn{4}{|c|}{LR-1} & \multicolumn{4}{c|}{LR-2} & \multicolumn{4}{c|}{LR-3} & \multicolumn{4}{c|}{LR-4} & \multicolumn{4}{c}{RL} \\ \midrule
% \multicolumn{1}{c|}{} & N & O & R & G & N & O & R & G & N & O & R & G & N & O & R & G & N & O & R & G \\ \midrule
% \multicolumn{1}{r|}{NONE (N)} & 273 & 0 & 0 & 2 & 432 & 0 & 0 & 0 & 433 & 0 & 0 & 0 & 420 & 3 & 0 & 0 & 2988 & 29 & 0 & 144 \\
% \multicolumn{1}{r|}{OFF (O)} & 0 & 0 & 0 & 0 & 0 & 0 & 0 & 0 & 0 & 0 & 0 & 0 & 0 & 1 & 0 & 0 & 0 & 0 & 0 & 0 \\
% \multicolumn{1}{r|}{RED (R)} & 0 & 42 & 1294 & 0 & 0 & 0 & 467 & 1 & 1 & 2 & 914 & 0 & 0 & 63 & 296 & 0 & 10 & 37 & 966 & 12 \\
% \multicolumn{1}{r|}{GREEN (G)} & 0 & 0 & 0 & 128 & 1 & 0 & 0 & 388 & 3 & 0 & 0 & 156 & 1 & 5 & 0 & 351 & 10 & 159 & 1 & 1753 \\ \bottomrule
% \end{tabular}
% \end{table*}

\begin{table*}[!t]
  \caption{Confusion Matrices for Each Testing Log ($\tau = 0.2$)}
  \label{tab:confusion-matrices}
  \centering
\renewcommand\tabcolsep{5.5pt}
\begin{tabular}{@{}rcccc|cccc|cccc|cccc|cccc@{}}
\toprule
 & \multicolumn{4}{|c|}{LR-1} & \multicolumn{4}{c|}{LR-2} & \multicolumn{4}{c|}{LR-3} & \multicolumn{4}{c|}{LR-4} & \multicolumn{4}{c}{RL} \\ \midrule
\multicolumn{1}{c|}{}          & N   & R    & G   & O  & N   &   R &   G & O &   N &   R &   G & O &   N &   R &   G &  O &    N &   R &    G &   O \\ \midrule
\multicolumn{1}{r|}{NONE (N)}  & 273 &    0 &   2 &  0 & 432 &   0 &   0 & 0 & 433 &   0 &   0 & 0 & 420 &   0 &   0 &  3 & 2988 &   0 &  144 &  29 \\
\multicolumn{1}{r|}{RED (R)}   &   0 & 1294 &   0 & 42 &   0 & 467 &   1 & 0 &   1 & 914 &   0 & 2 &   0 & 296 &   0 & 63 &   10 & 966 &   12 &  37 \\
\multicolumn{1}{r|}{GREEN (G)} &   0 &    0 & 128 &  0 &   1 &   0 & 388 & 0 &   3 &   0 & 156 & 0 &   1 &   0 & 351 &  5 &   10 &   1 & 1753 & 159 \\
\multicolumn{1}{r|}{OFF (O)}   &   0 &    0 &   0 &  0 &   0 &   0 &   0 & 0 &   0 &   0 &   0 & 0 &   0 &   0 &   0 &  1 &    0 &   0 &    0 &   0 \\ \bottomrule
\end{tabular}
\end{table*}

% 273 &    0 &   2 &  0 & 432 &   0 &   0 & 0 & 433 &   0 &   0 & 0 & 420 &   0 &   0 &  3 & 2988 &   0 &  144 &  29 \\
%   0 & 1294 &   0 & 42 &   0 & 467 &   1 & 0 &   1 & 914 &   0 & 2 &   0 & 296 &   0 & 63 &   10 & 966 &   12 &  37 \\
%   0 &    0 & 128 &  0 &   1 &   0 & 388 & 0 &   3 &   0 & 156 & 0 &   1 &   0 & 351 &  5 &   10 &   1 & 1753 & 159 \\
%   0 &    0 &   0 &  0 &   0 &   0 &   0 & 0 &   0 &   0 &   0 & 0 &   0 &   0 &   0 &  1 &    0 &   0 &    0 &   0 \\ \bottomrule

% \lcp{From Fig.~\lcp{X}, it is possible to notice that the system can recognize green traffic lights very early, but not red traffic lights.} This happens because, even though it knows where the traffic light is, no red TL bounding boxes fall within the defined threshold, so it predicts the TL is off. This is not necessarily harmful, since the car could be instructed to reduce to a stop on off and red TLs, and only proceed on green TLs, for safety. Taking this into consideration, a confusion matrix \lcp{where red, off, and yellow are a single class} is exhibited on Table~\ref{tab:confusion-matrices}, showing the system's performance on each testing log.

For qualitative analysis, a video for the system running can be seen at \url{https://youtu.be/VhdLpuErJ8E}; and, the code and trained model can be found at \url{https://github.com/LCAD-UFES/carmen_lcad/blob/master/src/traffic_light_yolo/README.md}.

\section{Conclusion}
\label{conclusion}

In this paper, we proposed a system for autonomous vehicles that uses deep learning and prior maps for traffic light recognition. First, a deep learning model performs traffic light detection and classification of state in a single step. Subsequently, prior maps are used to select only relevant traffic lights from the proposed detections, filtering out false positives as well. Additionally, a new method for creating prior maps is proposed. LiDAR points are projected to camera; points that hit inside detected bounding boxes are accumulated; and finally this point cloud is clustered to propose traffic lights' position in world coordinates.
% \lcp{(TODO) Sumarizar os resultados.}

Even though results are promising, much work should be done until such a system can be put on the roads. Particularly, it is important to improve the detector's performance in order to obtain more reliable results.

% Trabalhos futuros
% An alternative approach that we would like to explore in the future follows. Instead of detecting all traffic lights in an image and later filtering it, we could propose ROIs from prior maps (which is common in literature), and feed them to the deep learning model. This has the advantage that we do not have to run the detector in the entire image; providing opportunities for saving computational resources,
% or increasing accuracy, while maintaining similar computational cost, by feeding higher resolution segments.

As future work, we will investigate the reduction of the search space for traffic lights
by proposing ROIs in the input image based on prior maps annotations (which is common in literature). This approach has two main potential advantages: saving processing time by keeping the input image resolution as is, or, alternatively, enabling the use of higher resolution images in order to improve the system's performance.

% an alternative approach that we would like to explore in the future follows. Instead of detecting all traffic lights in an image and later filtering it, we could propose ROIs from prior maps (which is common in literature), and feed them to the deep learning model.
% This has the advantage of not running the detector on the entire image, thus providing an opportunity for saving computational resources.
%   % A:
% Or, if higher resolution images were used, an opportunity for improving the system's performance.
%   % B:
% Alternatively, this allows the use of higher resolution images which can improve the system's performance.

\section*{Acknowledgment}

The authors thank the NVIDIA Corporation for their kind donation of the GPUs used in this research.

\bibliography{references}

\begin{thebibliography}{10}

\bibitem{de2009traffic}
R.~De~Charette and F.~Nashashibi, ``{Traffic Light Recognition using Image
  Processing Compared to Learning Processes},'' in {\em International
  Conference on Intelligent Robots and Systems (IROS)}, 2009.

\bibitem{de2009real}
R.~De~Charette and F.~Nashashibi, ``{Real Time Visual Traffic Lights
  Recognition Based on Spot Light Detection and Adaptive Traffic Lights
  Templates},'' in {\em Intelligent Vehicles Symposium (IV)}, 2009.

\bibitem{siogkas2012traffic}
G.~Siogkas, E.~Skodras, and E.~Dermatas, ``{Traffic Lights Detection in Adverse
  Conditions using Color, Symmetry and Spatiotemporal Information},'' in {\em
  International Conference on Computer Vision Theory and Applications
  (VISAPP)}, 2012.

\bibitem{philipsen2015traffic}
M.~P. Philipsen, M.~B. Jensen, A.~M{\o}gelmose, T.~B. Moeslund, and M.~M.
  Trivedi, ``{Traffic Light Detection: A Learning Algorithm and Evaluations on
  Challenging Dataset},'' in {\em International Conference on Intelligent
  Transportation Systems (ITSC)}, 2015.

\bibitem{fairfield2011traffic}
N.~Fairfield and C.~Urmson, ``{Traffic Light Mapping and Detection},'' in {\em
  International Conference on Robotics and Automation (ICRA)}, 2011.

\bibitem{barnes2015exploiting}
D.~Barnes, W.~Maddern, and I.~Posner, ``{Exploiting 3D Semantic Scene Priors
  for Online Traffic Light Interpretation},'' in {\em Intelligent Vehicles
  Symposium (IV)}, 2015.

\bibitem{jensen2016vision}
M.~B. Jensen, M.~P. Philipsen, A.~M{\o}gelmose, T.~B. Moeslund, and M.~M.
  Trivedi, ``{Vision for Looking at Traffic Lights: Issues, Survey, and
  Perspectives},'' {\em IEEE Transactions on Intelligent Transportation
  Systems}, vol.~17, no.~7, pp.~1800--1815, 2016.

\bibitem{berriel2017grsl}
R.~F. Berriel, A.~T. Lopes, A.~F. de~Souza, and T.~Oliveira-Santos, ``{Deep
  Learning Based Large-Scale Automatic Satellite Crosswalk Classification},''
  {\em IEEE Geoscience and Remote Sensing Letters}, vol.~14, pp.~1513--1517,
  Sept 2017.

\bibitem{Berriel2017cag}
R.~F. Berriel, F.~S. Rossi, A.~F. de~Souza, and T.~Oliveira-Santos,
  ``{Automatic Large-Scale Data Acquisition via Crowdsourcing for Crosswalk
  Classification: A Deep Learning Approach},'' {\em Computers \& Graphics},
  vol.~68, pp.~32--42, Nov 2017.

\bibitem{GuidoliniIJCNN2018}
R.~Guidolini, L.~G. Scart, L.~F.~R. Jesus, V.~B. Cardoso, C.~Badue, and
  T.~Oliveira-Santos, ``{Handling Pedestrians in Crosswalks Using Deep Neural
  Networks in the IARA Autonomous Car},'' in {\em 2018 International Joint
  Conference on Neural Networks (IJCNN)}, pp.~1--8, July 2018.

\bibitem{redmon2016you}
J.~Redmon, S.~Divvala, R.~Girshick, and A.~Farhadi, ``{You only look once:
  Unified, real-time object detection},'' in {\em Proceedings of the IEEE
  conference on computer vision and pattern recognition}, pp.~779--788, 2016.

\bibitem{ren2015faster}
S.~Ren, K.~He, R.~Girshick, and J.~Sun, ``{Faster R-CNN: Towards real-time
  object detection with region proposal networks},'' in {\em Advances in neural
  information processing systems}, pp.~91--99, 2015.

\bibitem{weber2016deeptlr}
M.~Weber, P.~Wolf, and J.~M. Z{\"o}llner, ``{DeepTLR: A single Deep
  Convolutional Network for Detection and Classification of Traffic Lights},''
  in {\em Intelligent Vehicles Symposium (IV)}, 2016.

\bibitem{behrendt2017deep}
K.~Behrendt, L.~Novak, and R.~Botros, ``{A Deep Learning Approach to Traffic
  Lights: Detection, Tracking, and Classification},'' in {\em International
  Conference on Robotics and Automation (ICRA)}, 2017.

\bibitem{bach2018deep}
M.~Bach, D.~Stumper, and K.~Dietmayer, ``{Deep Convolutional Traffic Light
  Recognition for Automated Driving},'' in {\em International Conference on
  Intelligent Transportation Systems (ITSC)}, 2018.

\bibitem{pon2018hierarchical}
A.~D. Pon, O.~Andrienko, A.~Harakeh, and S.~L. Waslander, ``A hierarchical deep
  architecture and mini-batch selection method for joint traffic sign and light
  detection,'' {\em arXiv preprint arXiv:1806.07987}, 2018.

\bibitem{john2014traffic}
V.~John, K.~Yoneda, B.~Qi, Z.~Liu, and S.~Mita, ``{Traffic Light Recognition in
  Varying Illumination using Deep Learning and Saliency Map},'' in {\em
  International Conference on Intelligent Transportation Systems (ITSC)}, 2014.

\bibitem{baduesurvey}
C.~Badue {\em et~al.}, ``{Self-Driving Cars: A Survey},'' {\em arXiv preprint
  arXiv:1901.04407}, 2019.

\bibitem{diaz2015robust}
M.~Diaz-Cabrera, P.~Cerri, and P.~Medici, ``{Robust real-time traffic light
  detection and distance estimation using a single camera},'' {\em Expert
  Systems with Applications}, vol.~42, no.~8, pp.~3911--3923, 2015.

\bibitem{lindner2004robust}
F.~Lindner, U.~Kressel, and S.~Kaelberer, ``{Robust Recognition of Traffic
  Signals},'' in {\em Intelligent Vehicles Symposium (IVS)}, 2004.

\bibitem{levinson2011traffic}
J.~Levinson, J.~Askeland, J.~Dolson, and S.~Thrun, ``{Traffic Light Mapping,
  Localization, and State Detection for Autonomous Vehicles},'' in {\em
  International Conference on Robotics and Automation (ICRA)}, 2011.

\bibitem{franke2013making}
U.~Franke, D.~Pfeiffer, C.~Rabe, C.~Knoeppel, M.~Enzweiler, F.~Stein, and
  R.~Herrtwich, ``{Making Bertha See},'' in {\em International Conference on
  Computer Vision (ICCV) Workshops}, 2013.

\bibitem{jang2017traffic}
C.~Jang, S.~Cho, S.~Jeong, J.~K. Suhr, H.~G. Jung, and M.~Sunwoo, ``{Traffic
  light recognition exploiting map and localization at every stage},'' {\em
  Expert Systems With Applications}, vol.~88, pp.~290--304, 2017.

\bibitem{jensen2017evaluating}
M.~B. Jensen, K.~Nasrollahi, and T.~B. Moeslund, ``{Evaluating State-of-the-art
  Object Detector on Challenging Traffic Light Data},'' in {\em Conference on
  Computer Vision and Pattern Recognition Workshops (CVPRW)}, pp.~882--888,
  2017.

\bibitem{montemerlo2003perspectives}
M.~Montemerlo, N.~Roy, and S.~Thrun, ``{Perspectives on standardization in
  mobile robot programming: The Carnegie Mellon navigation (CARMEN) toolkit},''
  in {\em International Conference on Intelligent Robots and Systems (IROS)},
  2003.

\bibitem{de2016light}
L.~de~Paula~Veronese, J.~Guivant, F.~A.~A. Cheein, T.~Oliveira-Santos, F.~Mutz,
  E.~de~Aguiar, C.~Badue, and A.~F. De~Souza, ``{A Light-Weight Yet Accurate
  Localization System for Autonomous Cars in Large-Scale and Complex
  Environments},'' in {\em International Conference on Intelligent
  Transportation Systems (ITSC)}, 2016.

\bibitem{Mutz:2016:LMC:2873073.2873265}
F.~Mutz, L.~P. Veronese, T.~Oliveira-Santos, E.~de~Aguiar, F.~A. Auat~Cheein,
  and A.~Ferreira De~Souza, ``Large-scale mapping in complex field scenarios
  using an autonomous car,'' {\em Expert Syst. Appl.}, vol.~46, pp.~439--462,
  Mar. 2016.

\bibitem{ester1996density}
M.~Ester, H.-P. Kriegel, J.~Sander, X.~Xu, {\em et~al.}, ``{A Density-Based
  Algorithm for Discovering Clusters in Large Spatial Databases with Noise},''
  in {\em International Conference on Knowledge Discovery and Data Mining
  (KDD)}, 1996.

\bibitem{fregin2018driveu}
A.~Fregin, J.~M{\"u}ller, U.~Kre{\ss}el, and K.~Diermayer, ``{The DriveU
  traffic light dataset: Introduction and comparison with existing datasets},''
  in {\em International Conference on Robotics and Automation (ICRA)}, 2018.

\bibitem{everingham2010pascal}
M.~Everingham, L.~Van~Gool, C.~K. Williams, J.~Winn, and A.~Zisserman, ``{The
  Pascal Visual Object Classes (VOC) Challenge},'' {\em International Journal
  of Computer Vision}, vol.~88, no.~2, pp.~303--338, 2010.

\bibitem{lin2014microsoft}
T.-Y. Lin, M.~Maire, S.~Belongie, J.~Hays, P.~Perona, D.~Ramanan,
  P.~Doll{\'a}r, and C.~L. Zitnick, ``{Microsoft COCO: Common Objects in
  Context},'' in {\em European Conference on Computer Vision (ECCV)}, 2014.

\bibitem{yolov3}
J.~Redmon and A.~Farhadi, ``{YOLOv3: An Incremental Improvement},'' {\em arXiv
  preprint arXiv:1804.02767}, 2018.

\end{thebibliography}
\bibliographystyle{ieeetr}

\end{document}